%% file: 0_main.tex
\documentclass{llncs}
\usepackage{makeidx}  

\usepackage{amssymb,amsfonts,amsmath}
\usepackage{graphicx}
\usepackage{color}
\usepackage{float}
\usepackage{wrapfig}
\usepackage{amssymb}
\restylefloat{figure}
\usepackage{array}
\usepackage{multirow}
\usepackage[colorlinks]{hyperref}

\newcolumntype{P}[1]{>{\centering\arraybackslash}p{#1}}

\usepackage{url}
\urldef{\mailsa}\path|{marc.bolanos,petia.ivanova}@ub.edu|
\urldef{\mailsb}\path|{lvapeab,fcn}@prhlt.upv.es|

\DeclareMathOperator*{\argmax}{arg\,max~}

\newcommand{\q}{\mathbf{q}}

\newcommand{\h}{\mathbf{h}}

\newcommand{\E}{\mathbf{E}}
\newcommand{\U}{\mathbf{U}}

\newcommand{\W}{\mathbf{W}}

\newcommand{\R}{\mathbb{R}} 

\newcommand{\keywords}[1]{\par\addvspace\baselineskip
\noindent\keywordname\enspace\ignorespaces#1}

\usepackage{footnote}
\makesavenoteenv{tabular}
\usepackage[nameinlink,capitalize]{cleveref} 

\crefformat{appendix}{#2#1#3}

\begin{document}
\frontmatter          
\pagestyle{headings}  
\title{VIBIKNet: Visual Bidirectional Kernelized Network for Visual Question Answering}
\titlerunning{VIBIKNet}  
\author{Marc Bola\~nos\inst{1,2}, \'{A}lvaro Peris\inst{3}, Francisco Casacuberta\inst{3}, Petia Radeva\inst{1,2}}
\authorrunning{Marc Bola\~nos et al.} 
%
\tocauthor{Marc Bola\~nos, \'{A}lvaro Peris, Francisco Casacuberta, Petia Radeva}
\institute{Universitat de Barcelona, Barcelona, Spain,\\
	\email{marc.bolanos@ub.edu}, \email{petia.ivanova@ub.edu}
	\and
	Computer Vision Center, Bellaterra, Spain,\\
    \and
	PRHLT Research Center, Universitat Polit\`{e}cnica de Val\`{e}ncia, Val\`{e}ncia, Spain,\\
    \email{lvapeab@prhlt.upv.es}, \email{fcn@prhlt.upv.es}}

\maketitle              

\begin{abstract}

In this paper, we address the problem of visual question answering by proposing a novel model, called VIBIKNet. Our model is based on integrating Kernelized Convolutional Neural Networks and Long-Short Term Memory units to generate an answer given a question about an image. We prove that VIBIKNet is an optimal trade-off between accuracy and computational load, in terms of memory and time consumption. We validate our method on the VQA challenge dataset and compare it to the top performing methods in order to illustrate its performance and speed.

\end{abstract}

\input{1_introduction}
\input{3_methodology}
\input{4_results}
\input{5_discussion}

\subsubsection*{Acknowledgments.} This work was partially funded by TIN2015-66951-C2-1-R, SGR 1219, CERCA Programme / Generalitat de Catalunya,  CoMUN-HaT - TIN2015-70924-C2-1-R (MINECO/FEDER), PrometeoII/2014/030 and R-MIPRCV. P. Radeva is partially supported by ICREA Academia’2014. We acknowledge NVIDIA Corporation for the donation of a GPU used in this work.

\vspace{-0.5cm}

\bibliographystyle{abbrv}
\bibliography{0_main}

\end{document}

%% file: 1_introduction.tex
\section{Introduction}

Deep learning has proven to be applicable to several problems and data modalities (e.g. object detection, speech recognition, machine translation, etc.). Furthermore, it has been able to set new records, beating the state of the art in several artificial intelligence areas. Now, new machine learning problems may be tackled, taking profit from the capabilities of deep learning methods for combining multiple data modalities and be end-to-end trainable, thus, having potential to enable new research and application areas. Some multimodal problems are image captioning~\cite{xu2015show}, video captioning~\cite{peris2016video} or multimodal machine translation and crosslingual image captioning~\cite{Specia16}. In this work, we address the challenging Visual Question Answering (VQA)~\cite{Antol15} problem.

\input{2_related_work}

In this work, we propose a model for open-ended VQA which uses the most powerful state-of-the-art methods for image and text characterization. More precisely, we use a Kernelized CNN (KCNN) for image characterization, which takes profit from detecting and characterizing all objects in the image for generating a combined feature descriptor. For question modeling, we apply pre-trained word embeddings from Glove~\cite{Pennington14}, taking advantage from the transfer learning capabilities of neural networks; and a Bidirectional LSTM (BLSTM), able to learn rich question information by taking into account temporal relationships both in past-to-future and future-to-past manner. Next, we fuse both modalities and finish by applying a classification model for obtaining the resulting answer.

This paper is organized as follows: in \cref{sec:methodology}, we present the proposed method, VIBIKNet. In \cref{sec:results}, we describe the dataset and the evaluation metrics used. We evaluate our model and compare it with the state of the art. Finally, in \cref{sec:conclusions}, we give some concluding remarks and some directions of future work.

%% file: 2_related_work.tex
From the visual modality perspective, a clear proposal for processing images are Convolutional Neural Networks (CNNs)~\cite{szegedy2015going}. CNNs are a powerful tool, not only for image classification, 
but also for feature extraction. 
Nevertheless, they are not fully scale and rotation invariant, unless they have been specifically trained with enough varied examples~\cite{cheng2016rifd}. Furthermore, this invariance problem gets more acute in scene images, which are composed of multiple elements at possibly different rotations and scales. In order to tackle this problem, Liu proposed in \cite{liu2015kernelized} a Kernelized approach for learning a rich representation for images composed of multiple objects in any possible rotation and scale.

From the textual modality perspective, Recurrent Neural Networks (RNNs) have shown to be effective sequence modelers. The use of gated units, such as Long Short-Term Memory (LSTM)~\cite{Hochreiter97}, allows to properly process long sequences. In the last years, LSTM networks have been used in a wide variety of tasks, such as machine translation~\cite{Sutskever14} or image and video captioning~\cite{xu2015show,peris2016video}.

After the appearance of the VQA dataset~\cite{Antol15} and the organization of the VQA Challenge, several models appeared addressing this problem. Some notable examples are the ones by Kim et al. \cite{kim2016multimodal}, where image and question was separately described by a CNN and by a RNN, and then a Multimodal Residual Network (MRN) was used for combining both modalities. Fukui et al. \cite{fukui2016multimodal} used a CNN for describing the image and a two-layered LSTM for the question; followed by a Multimodal Compact Bilinear Pooling (MCB) for fusion. Nam et al. \cite{nam2016dual}, after describing the input image and question, applied a powerful Dual Attention Network (DAN) for fusing both modalities.

%% file: 3_methodology.tex
\section{VIBIKNet} \label{sec:methodology}

\begin{figure}[ht!]
\vspace{-0.4cm}
\centering
\includegraphics[width=0.7\textwidth]{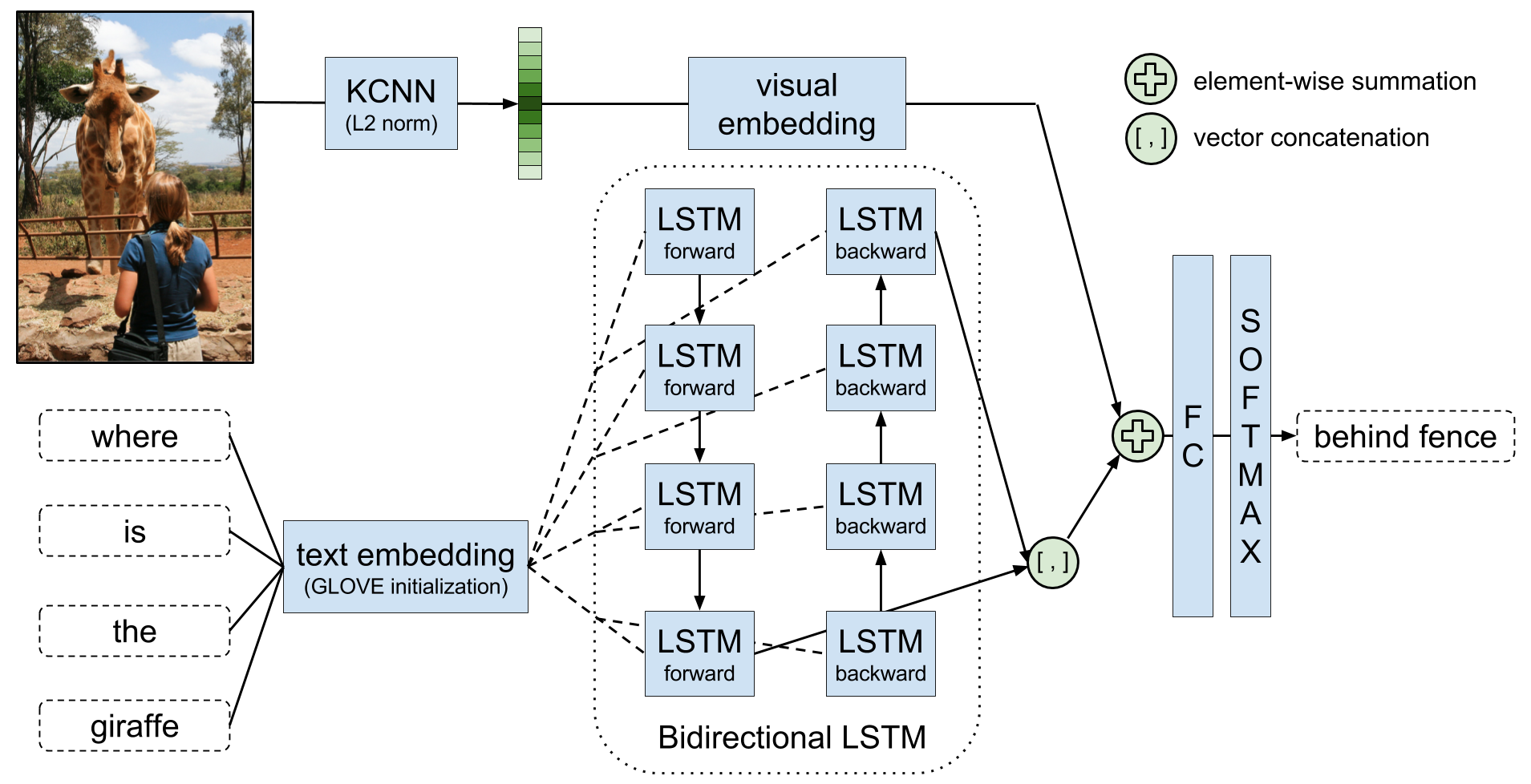}
\caption{General scheme of the proposed VIBIKNet model.}
\vspace{-0.3cm}
\label{fig:model_VIBIKNet}
\end{figure}

In this section, we describe our VQA system, named Visual Bidirectional Kernelized Network (VIBIKNet), whose general scheme can be seen in~\cref{fig:model_VIBIKNet}. We also make public the complete source code\footnote{\url{https://github.com/MarcBS/VIBIKNet}} for reproducing the results obtained.

The VQA problem consists in computing a function $f$ which, having as input an image $X$ and a related question $Q$, produces a textual answer $A$:

\vspace{-1em}
\begin{equation}
	f(X,Q) = A
\end{equation}
where $Q$ and $A$ are two variable-sized sequences of words, which can be formalized as $Q = q_1, q_2, ..., q_N$ and $A = a_1, a_2, ..., a_M$, respectively.

We formulate the problem under a probabilistic framework. Given the clear multimodality of it, first, we propose to extract independent representations for image and question. For obtaining a rich representation of the image, we apply a KCNN~\cite{liu2015kernelized} (\cref{sec:kcnn}). 
We process the question with a BLSTM network (\cref{sec:BLSTM}), which considers the full question context. 
Next, we need to combine modalities into a single representation. To this purpose we propose using a simple, yet effective, element-wise summation 
(see~\cref{sec:multimodal_fusion}) after embedding the visual information into the textual one. Finally, we predict the output answer, which can be estimated with a simple classifier for the dataset at hand.

\vspace{-1cm}

\subsection{KCNN for image representation} 
\label{sec:kcnn}

A key factor that makes humans able to understand what happens on a picture is the ability to distinguish each of the present elements in it, regarding any possible scale or orientation, together with the relationships and actions that are taking place between them. When we talk about elements we refer to any object, person or animal appearing in the images.

\begin{figure}[ht!]
\vspace{-0.3cm}
\centering
\includegraphics[width=0.7\textwidth]{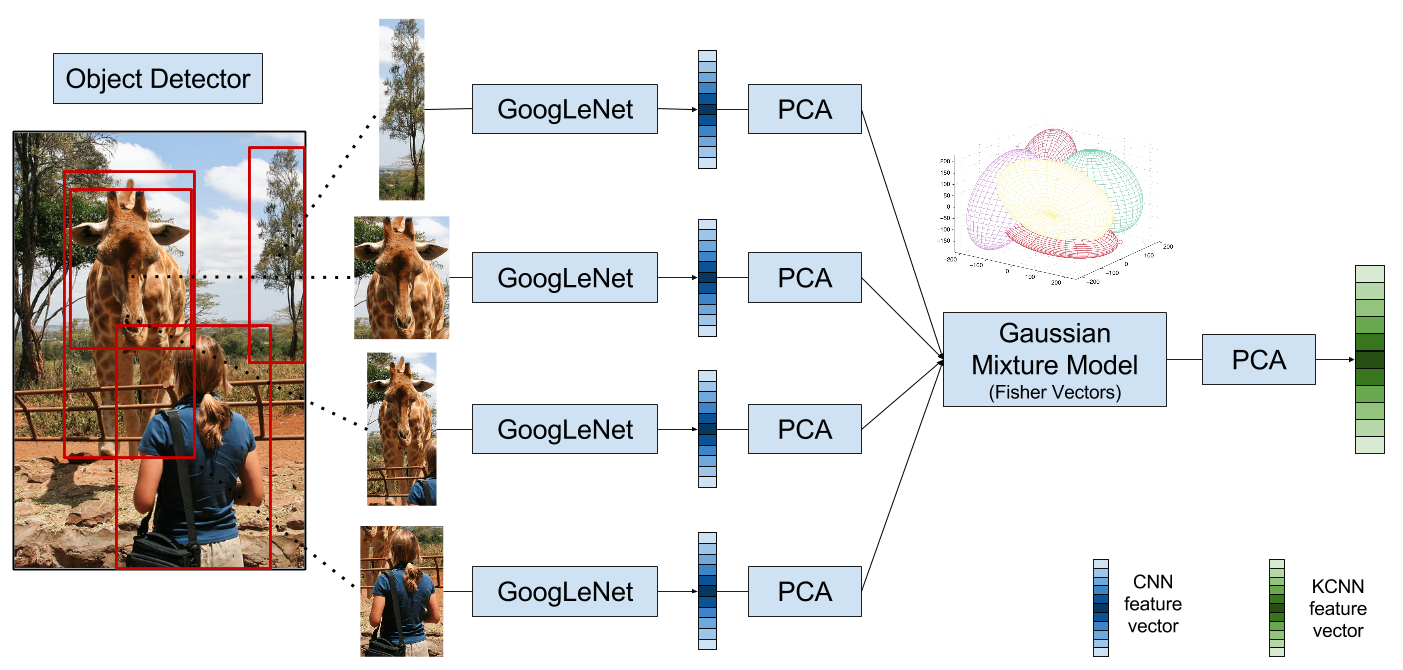}
\caption{Steps for the extraction of Kernelized CNN features.}
\label{fig:KCNN}
\vspace{-0.2cm}
\end{figure}

Following this idea, the so-called Kernelized Deep Convolutional Neural Network method~\cite{liu2015kernelized} has the ability to capture all these aspects. In~\cref{fig:KCNN} we show the general pipeline of steps for extracting KCNN features from images. 

More formally, given two images, $X$ and $Y$, and a set of variable-sized regions for each of them $X = \{x_1, x_2, ..., x_n\}$, and $Y = \{y_1, y_2, ..., y_m\}$, we can define their similarity given by a kernel $K$ as:
%
\begin{equation} \label{eq:kernel2}
    K(X,Y)  = \ < \sum_{x_i \in X} \psi(x_i) , \sum_{y_j \in Y} \psi(y_j) > \\
            = \ < \Psi(X) , \Psi(Y) >
\end{equation}
where the similarity between two regions is computed by their inner product, $\psi$ denotes a linear/non-linear transformation and $\Psi$ denotes the final vectorial image representation composed by the set of initial regions.

Going back to the general scheme applied, initially, an object detector is used for extracting object candidate bounding boxes from each image, $x_i$. After that, and in order to provide robustness to the point of view, a set of rotations are applied separately to each of the extracted image regions before extracting their image features through a CNN, $\psi$ in \cref{eq:kernel2}. Next, a PCA transformation is applied to the vectors from all image regions. In order to aggregate all vectors from a single image, we learn a Fisher kernel \cite{perronnin2010improving} which, similarly to a Bag-of-Words approach \cite{sivic2009efficient}, jointly models the features distribution by learning a Gaussian Mixture Model (GMM), namely $\Psi$ in \cref{eq:kernel2}. In order to have manageable vector sizes, an additional PCA is applied to the resulting aggregated vectors. This produces an $l$-size representation of the image, which is finally normalized in order to obtain the final representation of the image ($\Phi(X)$).

\subsection{Bidirectional LSTM for question representation} \label{sec:BLSTM}

As stated above, a question $Q = q_1, q_2, ..., q_N$ is a variable-sized sequence of words. We use a powerful sequence modeler such a RNN for characterizing $Q$: each word is inputted to the system following a $1$-hot codification. Next, we project each word to a continuous space by means of a learnable word embedding matrix. In order to effectively train our word embedding model, we start from pre-trained word vectors provided by Glove~\cite{Pennington14} and we fine-tune them with the questions corpus. Words not included in Glove are randomly initialized. 

The sequence of word embeddings 
is then inputted to a bidirectional RNN. Bidirectional RNNs are made up of two independent recurrent layers, each of them analyzing the input sequence in one direction. Hence, the forward layer processes the sequence from the left to the right while the backward layer process it from the right to the left. In our case, each recurrent layer is an LSTM layer.

LSTM networks allow to deal with the vanishing gradient problem. These layers maintain two internal states, namely the hidden state ($\h$) and the memory state ($\bf c$). The amount of information that flows through the network is modulated by the input (${\bf i}$), output (${\bf o}$) and forget (${\bf f}$) gates. Refer to~\cite{Gers00} for a more in-depth review of the LSTM networks.

For obtaining a representation of the complete question, we concatenate the last hidden state from the forward and backward layers:

\begin{equation}
\mathbb{Q} = [\h_N^{\mathrm f}, \h_N^{\mathrm b}]
\label{Eq:1}
\end{equation}
where $\h_N^{\mathrm f} \in \mathbb{R}^m $ and $\h_N^{\mathrm b} \in \mathbb{R}^m$ are the last forward and backward hidden states, of size $m$. $[\cdot ~,~ \cdot]$ denotes vectorial concatenation and $\mathbb{Q}$ is the final representation of $Q$. Since each LSTM layer processes the complete input sequence in one direction, $\mathbb{Q}$ contains both left-to-right and right-to-left dependencies.

\subsection{Multimodal fusion and prediction} \label{sec:multimodal_fusion}

\textbf{Multimodal fusion.}
Our problem involves information from two different sources. Hence, we must combine both image and text, given that image $X$ is represented by a KCNN as a feature vector $\Phi(X)$ of size $l$, and that question $Q$ is represented as $\mathbb{Q}$, of size $2m$.

In order to properly combine both modalities, we first linearly project the image representation to the same space as the question representation, by means of a \textit{visual embedding} matrix:
\begin{equation}
    \mathbb{X} = \W_m \Phi(X)
\end{equation}
where $\W_m$ is a $2m \times l$ matrix, jointly estimated with the rest of the model.

Then, a fusion operation is applied on both modalities, $\mathbb{X}$ and $\mathbb{Q}$:

\begin{equation}
\mathbb{M} = \mathbb{X} \oplus \mathbb{Q}
\label{eq:fusion}
\end{equation}
where $\oplus$ is the fusion operator and $\mathbb{M}$ is the joint, multimodal representation of the image and question.

\textbf{Prediction.}
Given the nature of the task at hand, a typical answer has few words. More precisely, in the VQA dataset (\cref{sec:dataset}), the 89.3\% of the answers are single-worded; and the 99.0\% of the answers have three or less words~\cite{Antol15}. 

Therefore, we treat our problem as a classification task over the $K$ most repeated answers. The obtained fusion of vision and text ($\mathbb{M}$) is inputted to a fully-connected layer with the set of answers as output. Applying a softmax activation, we define a probability over the possible answers. At test time, we choose the answer $\hat{a}$ with the highest probability:
\begin{equation}
\hat{a} = \argmax_{a\in K}p(a | Q, X)
\end{equation}

%% file: 4_results.tex
\section{Experiments and results} \label{sec:results}

In this section we set up the experimentation and evaluation procedure. Moreover, we study and discuss the obtained results in the VQA Challenge\footnote{The VQA Challenge leaderboard is available at \url{http://visualqa.org/roe.html}}.

\subsection{Dataset and evaluation}
\label{sec:dataset}
We evaluate our model on the VQA dataset~\cite{Antol15}, on the real open-ended task. The dataset consists of approximately 200,000 images from the MSCOCO dataset~\cite{chen2015microsoft}. Each image has three questions associated and each question has ten answers, which were provided by human annotators. We used the default splits for the task: \textit{Train} (80,000 images) for training, \textit{Test-Dev} (40,000 images) for validating the model and \textit{Test-Standard} (80,000 images) for testing it. An additional partition, \textit{Test-Challenge}, was used for evaluating the model at the VQA Challenge. 

We followed the VQA evaluation protocol~\cite{Antol15}, which computes an accuracy between the system output ($\hat{a}$) and the answers provided by the humans:
\begin{equation}
Acc (\hat{a}) = \min \Big\{ \frac{\textsf{{\#} humans that said }\hat{a}}{3}, 1 \Big\}
\end{equation}

\subsection{Experimental setup}

We set the model hyperparameters according to empirical results. For extracting the KCNN features, we used: EdgeBoxes \cite{ZitnickECCV14edgeBoxes} for proposing 100 object regions, a set of 8 different object rotations of $R = \{0, 45, 90, 135, 180, 225, 270, 315\}$ degrees, the last FC layer of GoogLeNet ~\cite{szegedy2015going} (1024-dimensional) for extracting features on each object, applied a PCA of dimensions 128 before, and $l=1024$ after the GMM, respectively, and learned 128 gaussians during GMM training. 

Since we used Glove vectors, the word embedding size was fixed to 300. The BLSTM network had $m=250$ units in each layer, which accounts for a total 500 units. The visual embedding had a size of $2m=500$. We applied a classification over the 2,000 most frequent answers, which covered a 86.8\% of the whole dataset. As fusion operator ($\oplus$ in~\cref{eq:fusion}) we tested element-wise summation and element-wise concatenation. Following~\cite{fukui2016multimodal}, we also tested MCB pooling as $\oplus$. 

We used the Adam~\cite{Kingma14} optimizer with an initial learning rate of $10^{-3}$. As regularization strategy, we only applied dropout before the classification layer. 

\subsection{Experimental results}\label{sec:exp_results}

\cref{tab:results} shows the accuracies of variations of our model (top) and of other works (bottom) for the \textit{Test-Dev} and \textit{Test-Standard} splits. Results are separated according to the type of answer, namely yes/no (Y/N), numerical (Num.) and other (Other) answers. We also report the overall accuracy of the task (All).
\begin{table}[th]
	\centering
\def\arraystretch{1}
\setlength\tabcolsep{5pt}
	\caption{Proposed models compared to the state of the art. G stands for GoogLeNet, R for ResNet-152, K for KCNN, L for LSTM, BL for BLSTM, FC for fully-connected layer on text before fusing, sum for fusion by summation, cat for fusion by concatenation, +val for training using train+val, and VIBIKNet is "G-K BL sum". Some results are unavailable (--).\label{tab:results}}
    \vspace{-0.2cm}
	\begin{tabular}{l c c c c c c c c}
		\hline
		 &  \multicolumn{4}{c}{\textbf{Test-Dev}~[\%]}  &  \multicolumn{4}{c}{\textbf{Test-Standard}~[\%]}\\	\cline{2-9}
		 & Y/N & Num. & Other & All & Y/N & Num. & Other & All \\ \hline
	    G-K L sum & 79.0 & 38.2&33.7 &52.9 & --& --&-- &--  \\ 
       G-K BL FC sum & 78.6& 33.6&36.9 & 52.1 & --& --&-- &--  \\ 
       G-K BL FC cat & 79.0 & 33.6& 38.3 & 53.0 & --& --&-- &--  \\
       R BL sum & 77.8 & 30.6 & 38.6 & 52.3 & -- & -- & -- & -- \\
       G-K BL cat & 79.0& 33.4&38.5 & 53.0 & --& --&-- &-- \\
       G-K BL MCB & 79.2& 33.2&37.5 & 52.5 & --& --&-- &-- \\       
       VIBIKNet & 79.1& 38.3&33.5 & 53.1&78.3&38.9 &39.0 &54.9 \\
	   VIBIKNet +val & --& --& --& -- &78.9 &36.3 & 40.3& 55.8\\  \hline

       MRN \cite{kim2016multimodal} & -- & -- & -- & -- & 82.4 & 38.2 & 49.4 & 61.8 \\
       DAN \cite{nam2016dual} & 83.0 & 39.1 & 53.9 & 64.3 & 82.8 & 38.1 & 54.0 & 64.2 \\
       MCB \cite{fukui2016multimodal} & 82.3 & 37.2 & 57.4 & 65.4 & -- & -- & -- & --  \\
       Human \cite{Antol15} & -- & -- & -- & -- & 95.8 & 83.4 & 72.7 & 83.3 \\
       \hline
	\end{tabular}
	\vspace{-0.5cm}
\end{table}

\begin{figure}[H]
\centering
\includegraphics[width= 
						.8\textwidth]{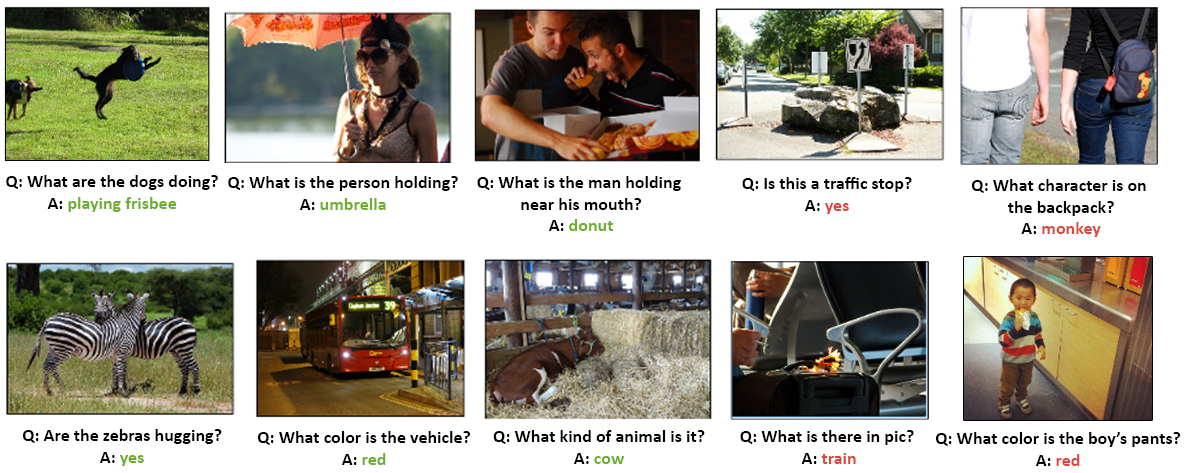}
\caption{Examples of the predictions provided by VIBIKNet; in green correctly predicted and in red wrongly predicted answers.}
\label{fig:VIBIKNet_examples}
\vspace{-0.3cm}
\end{figure}

It can be seen that both summation and concatenation fusion strategies performed similarly. In terms of performance, MCB was also similar to them. Nevertheless, MCB was much more resource-demanding: while the average time per epoch of summation was 320s, MCB required approximately 5,800s. Moreover, adding a fully-connected layer after text characterization and before fusion did not help, meaning that the visual embedding mechanism suffices for providing a robust visual-text embedding.
Regarding image characterization, if we compare the results using ResNet-152 vs GoogLeNet-KCNN, we can see that even using a less powerful CNN architecture, the adoption of the KCNN representation provided better results than simply using the ResNet output. Finally, it is worth noting that we used a single model for prediction. The use of network ensembles typically offer a performance boost~\cite{fukui2016multimodal}.  In~\cref{fig:VIBIKNet_examples} we can see some qualitative examples of our methodology.

\vspace{-0.5cm}
\vspace{-1cm}


%% file: 5_discussion.tex
\vspace{-0.4em}

\section{Conclusions and Future Work} \label{sec:conclusions}

We proposed a method for VQA which offers a trade-off between the accuracy and the computational cost of the model. We have proven that kernelized methods for image representation based on CNNs are very powerful for the problem at hand. Additionally, we have shown that using simple fusion methods like summation or concatenation can produce similar results to more elaborate methods at the same time that provide a very efficient computation. Nevertheless, we are aware that performing the multimodal fusion at deeper levels may be beneficial.

As future directions, we aim to delve into better fusion strategies but keeping a low computational cost. We extracted KCNN features based on local representations (objects appearance), but using them together with end-to-end trainable attention mechanisms may lead to higher performances~\cite{fukui2016multimodal}. 

\vspace{-0.5em}